\documentclass[man,floatsintext]{apa7}
\usepackage{lineno}   
\usepackage{lipsum}
\usepackage{caption}
\usepackage{bm}
\usepackage{makecell}
\usepackage{booktabs}
\usepackage{tabularx}
\usepackage[american]{babel}
\usepackage{enumitem}
\usepackage{wrapfig}
\usepackage{amsfonts}       
\usepackage{amsmath,amsfonts, amssymb}
\usepackage{csquotes}
\usepackage[style=apa,sortcites=true,sorting=ynt,backend=biber]{biblatex}
\DeclareLanguageMapping{american}{american-apa}

\title{Neural Simulation-Based Inference for Computational Cognitive Models: A Tutorial}
\shorttitle{Tutorial for Neural SBI}

\authorsnames[1,1,{1,2}]{Ti-Fen Pan, Yi Cheng, Anne GE Collins}
\authorsaffiliations{{Department of Psychology, University of California, Berkeley},{Helen Wills Neuroscience Institute, University of California, Berkeley}}

\leftheader{Pan}

\abstract{
Computational cognitive modeling is an increasingly essential methodology to elucidate the mechanisms underlying human and animal cognition. By fitting models to empirical data, researchers can perform parameter estimation, model comparison, and model-based neural analyses to assess competing theories, quantify individual differences, and investigate neural representations. Typically, fitting these models relies on likelihood-dependent methods (e.g., Maximum likelihood estimation or hierarchical Bayesian methods). However, for many reasonable cognitive models, computing the likelihood is computationally intractable, in which case these methods become inadequate. This substantially limits the range of theories that researchers can investigate. Simulation-based inference with neural networks (neural SBI) has emerged as a powerful toolkit to overcome this hurdle. By training deep neural networks on datasets generated through model simulations, neural SBI learns the relation between observable data and cognitive models. This approach enables all the applications of standard modeling while bypassing the likelihood computation, without constraints on empirical data size. Despite its successful adoption across diverse scientific fields, neural SBI remains underutilized in cognitive modeling, partly due to a lack of accessible, domain-tailored guidance. This tutorial addresses this gap by providing a clear workflow for applying neural SBI to cognitive models. We also provide practical use cases and recommendations for reducing the computational and simulation burden, facilitating efficient implementation in real-world research settings. By lowering the technical barriers to neural SBI, this tutorial aims to empower cognitive scientists to investigate richer theory spaces and ultimately propel scientific discovery.
}


\authornote{
  Corresponding author: \addORCIDlink{Ti-Fen Pan}{0000-0002-2640-7664}, Department of Psychology, University of California, Berkeley, 2121 Berkeley Way, 94704.  E-mail: tfpan@berkeley.edu}

\begin{document}
\maketitle
Computational cognitive models are mathematical tools designed to characterize the underlying computations that give rise to human behavior. By fitting these models to empirical data, researchers can test competing theories via model comparison, quantify individual differences through parameter estimation, or probe neural mechanisms by linking brain activity to inferred trial-by-trial latent dynamics. Traditionally, fitting cognitive models relies on likelihood-based estimation methods, such as Maximum Likelihood Estimation (MLE; \cite{myung2003tutorial}), Maximum a Posteriori (MAP; \cite{cousineau2013improving}) or hierarchical Bayesian modeling \parencite{baribault2023troubleshooting}. However, these methods become unsuitable when the likelihood function is analytically intractable (lacking a closed-form solution) or computationally intractable (requiring prohibitive computational time). These limitations constrain the range of cognitive models that researchers can practically assess, thereby limiting the range of hypotheses that can be formally tested about the mechanisms underlying human cognition.

\begin{figure}[t!]
    \centering
    \includegraphics[width=1\linewidth]{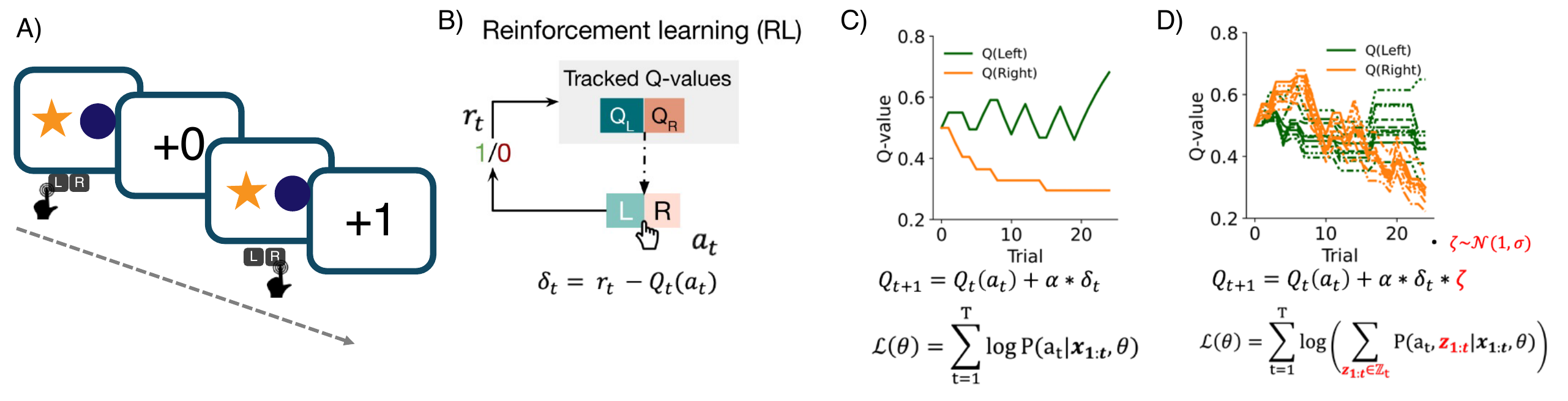} 
    \caption{\textbf{Intractable model example} A) Two-armed bandit task. B) A reinforcement learning (RL) agent tracks action values $Q_L/Q_R$ using the standard delta rule. At trial $t$, an agent selects an action $a_t$, receives feedback $r_t$, and updates the chosen action value $Q_t(a_t)$ proportionally to the reward prediction error $ \delta$. C) A noiseless RL model with a tractable likelihood $L(\theta)$. D) A noisy RL model with an intractable likelihood. Value updates are corrupted by Gaussian noise $\zeta$. Past observations are denoted by $x_{1:t}$, all possible past $Q_L/Q_R$ paths by $z_{1:t}$, and the parameters to be recovered (e.g., the learning rate $\alpha$) by $\theta$.}    
    \label{fig:sim_intractable_exp}
    \vspace{-25pt}
\end{figure}

Importantly, intractable likelihoods do not only arise in overly complex, edge-case models. While a likelihood can become intractable for various reasons, even simple, cognitively relevant examples illustrate the broad scope of this problem. For example, cognitive models that incorporate noisy sequential processes, either within trials \parencite{leng2026perils} or across them \parencite{findling2019computational}, add exponentially expanding uncertainty to the model variables. Computing the likelihood in these cases requires marginalizing out this uncertainty, which is computationally intractable (Fig.~\ref{fig:sim_intractable_exp}). Similarly, models that rely on unobservable latent states that condition observable data, such as hierarchical reinforcement learning \parencite{eckstein2020computational} or state-space models \parencite{ashwood2022mice}, suffer from the same intractability. These models can nonetheless represent relevant, useful cognitive theories. However, the challenge of fitting such models often forces researchers to rely on simplified, tractable alternatives that are significantly different, such as substituting the intractable leaky competing accumulator \parencite{usher2001time} with the tractable linear ballistic accumulator \parencite{brown2008simplest}; to develop ad-hoc statistical estimation techniques tailored for each specific problem \parencite{ashwood2022mice, findling2019computational}; or to abandon formal fitting techniques altogether \parencite{eckstein2020computational}.

Several methods exist to overcome the challenge of intractable likelihoods.  Researchers can approximate the likelihood or posterior by deriving model-specific mathematical approximations or using sampling-based methods such as particle filtering or Markov chain Monte Carlo (MCMC; \cite{gilks1995markov}). However, model-specific approximations require sophisticated mathematical expertise and are often difficult to generalize across models, whereas sampling-based methods can be computationally expensive, particularly when facing large datasets. The increasing availability of computational resources, together with rapid advances in machine learning, has therefore motivated an alternative approach: simulation-based inference (SBI).

SBI methods are likelihood-free techniques that infer the relationship between cognitive models and observed data using forward simulations generated directly by the models. While Approximate Bayesian Computation (ABC) served as the earliest accessible SBI framework in cognitive modeling \parencite{sunnaaker2013approximate}, it required researchers to hand-craft summary statistics and specify a distance metric for comparing simulated and empirical data. These requirements can become particularly restrictive for complex or high-dimensional data, where informative summary statistics may be difficult to identify. To overcome these limitations, modern deep learning has catalyzed the rise of \emph{neural SBI}, a family of methods that train artificial neural networks on simulated data to directly approximate posteriors (e.g., best-fitting parameters), likelihoods, or likelihood ratios \parencite{cranmer2020frontier, fengler2021likelihood, zammit2025neural}.

Neural SBI has been developed across many fields, including physics \parencite{dax2021real, regamey2026galaxy}, biology \parencite{dingeldein2025simulation, wang2026agent}, economics \parencite{shiono2021estimation, dyer2022black}, and neuroscience \parencite{ghaderi2023general, zeraati2026neural}, and has substantially broadened modeling or hypothesis landscapes. A growing body of work in cognitive science/psychology has similarly leveraged neural SBI to propel varied cognitive model development \parencite{fengler2021likelihood, sokratous2023ask, schumacher2025validation, rmus2024artificial, kvam2025comparing, pan2025latent, maher2026learning}. Yet, despite the promising avenues of neural SBI and the emergence of broad technical guidelines \parencite{deistler2025simulation, arruda2025diffusion}, a focused, domain-specific tutorial remains a barrier to broader adoption in cognitive modeling.  

To bridge this gap, this tutorial provides a step-by-step practical guide to neural SBI tailored specifically for cognitive modelers, including those without extensive machine learning backgrounds. In the sections that follow, we first introduce the foundations of neural SBI, outline its primary use cases in cognitive modeling, and provide concrete advice for when to adopt it over classical methods. Next, we walk through an end-to-end implementation workflow with practical recommendations across common modeling scenarios. Finally, we highlight accessible existing software, practical considerations, and current limitations when applying neural SBI to empirical behavioral data. 

\section{Simulation-based inference with neural networks}
\begin{figure}[t!]
    \centering
    \includegraphics[width=1\linewidth]{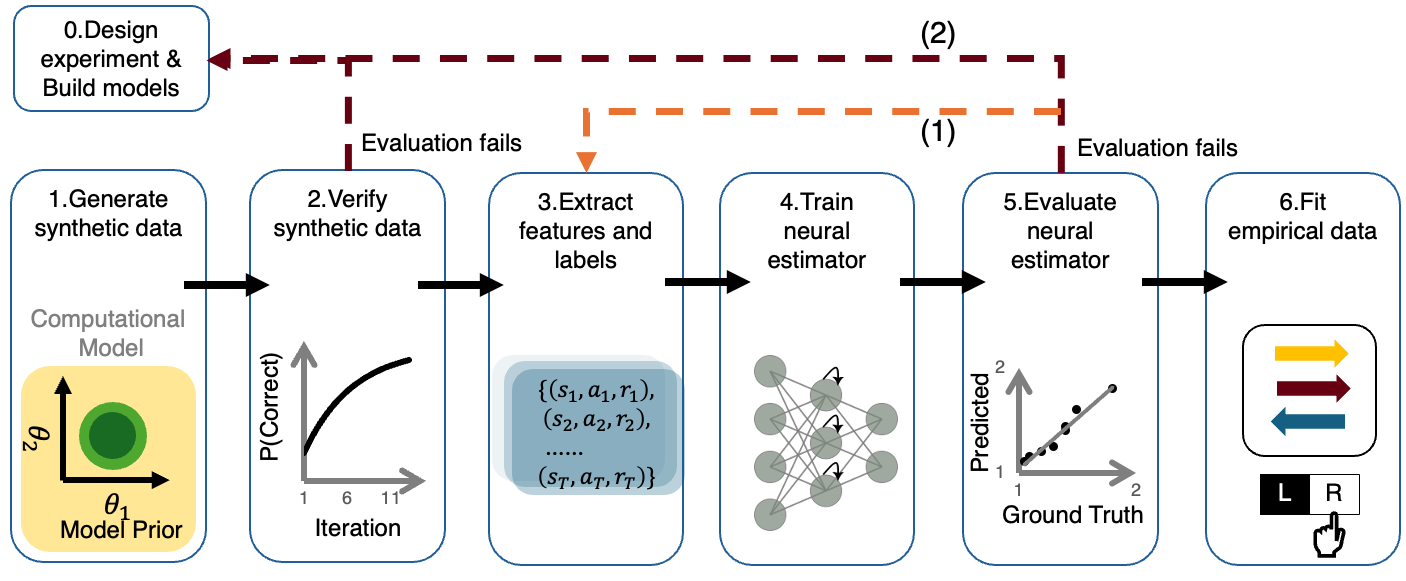}
    \vspace{-15pt}    
    \caption{\textbf{Workflow of neural SBI.} Steps 1- 5: A complete pipeline to train a neural estimator with a synthetic dataset generated from well-defined computational cognitive models and experiments. Step 6: A trained neural estimator is amortized to fit real data. Dashed lines denote the iterative process if the evaluation/validation fails.}    
    \label{fig:sbi_workflow}
    \vspace{-25pt}    
\end{figure}
\subsection{What is neural SBI?}
Simulation-based inference with neural networks (neural SBI) is a framework that uses artificial neural networks (ANNs) and simulated data from computational models (typically few-parameter cognitive models) to approximate the relationship between the computational models and observed data. Here, we refer to these trained ANNs as \textit{neural estimators}. Once trained on simulated data, these neural estimators can be applied on real participant data to \emph{infer} unknown computational model variables from behavior. Recent advances in neural estimators have led to several approaches, including neural posterior estimation (NPE), neural likelihood estimation (NLE), and neural ratio estimation (NRE), each with distinct advantages and limitations (see \cite{zammit2025neural} for a detailed comparison). The workflow presented here focuses on NPE and its point estimation variant, neural point estimation, as they represent the most accessible and adopted neural SBI techniques in cognitive modeling  \parencite{schumacher2025validation, schumacher2023neural, kvam2025comparing, rmus2024artificial, pan2025latent}. Nevertheless, the core principles of this tutorial generalize to other neural SBI methods. 

Neural SBI is grounded in the principle that most computational models, including those with intractable likelihoods, can be forward-simulated to generate synthetic data that matches the format of empirical behavior. Because the data is simulated, this synthetic data can be labeled with known ground-truth model identity, parameters, and latent variables. The workflow therefore begins by simulating synthetic data from candidate cognitive models under the exact experimental design (Fig.~\ref{fig:sbi_workflow} Step 1), creating a labeled dataset to train and validate the neural estimators. To ensure the validity of both synthetic data and cognitive models, we evaluate the simulated outputs against predefined, qualitative, and model-independent metrics that capture key behavioral phenomena (e.g., learning curves, Fig.~\ref{fig:sbi_workflow} Step 2; ~\cite{wilson2019ten}). This sanity check confirms the integrity of the simulation code and that the experimental design elicits the target behaviors assumed by the computational models. If this check fails, the experimental design or cognitive models must be revised before proceeding.

Next, the validated synthetic data are formatted into input training features and target labels (Fig.~\ref{fig:sbi_workflow} Step 3). Input features consist exclusively of observable experimental variables (e.g., stimuli, choices, reaction times) that can be collected from real participants. By contrast, target labels correspond to the latent cognitive variables of interest (e.g., learning rate) that we aim to infer from real participants. During the \textit{training phase} (or \textit{training time}), an untrained neural estimator iteratively updates its internal weights to learn the statistical mapping from observable features to target labels, through standard supervised training techniques (Fig.~\ref{fig:sbi_workflow} Step 4). The trained neural estimator is then evaluated on a held-out simulated test set to assess its generalizability and detect potential over- or underfitting (Fig.~\ref{fig:sbi_workflow} Step 5). Poor recovery at this stage indicates a need to revisit feature selection/extraction (Step 3) or experimental design (Step 0). Once validated, the neural estimator is ready for the \textit{inference phase} (or \textit{inference time}), during which its weights remain fixed as it processes unseen data to generate predictions. Specifically, real participant data are provided as input to infer the underlying cognitive variables (Fig.~\ref{fig:sbi_workflow} Step 6). Crucially, neural SBI enables \emph{amortized} inference: the computational cost is front-loaded during training, making subsequent inference computationally inexpensive. Consequently, a single trained network can perform rapid inference (e.g., model comparison, parameter estimation) across multiple empirical datasets without retraining or per-participant optimization, contrary to likelihood-based techniques.

\begin{figure}[t!]
    \centering
    \includegraphics[width=1\linewidth]{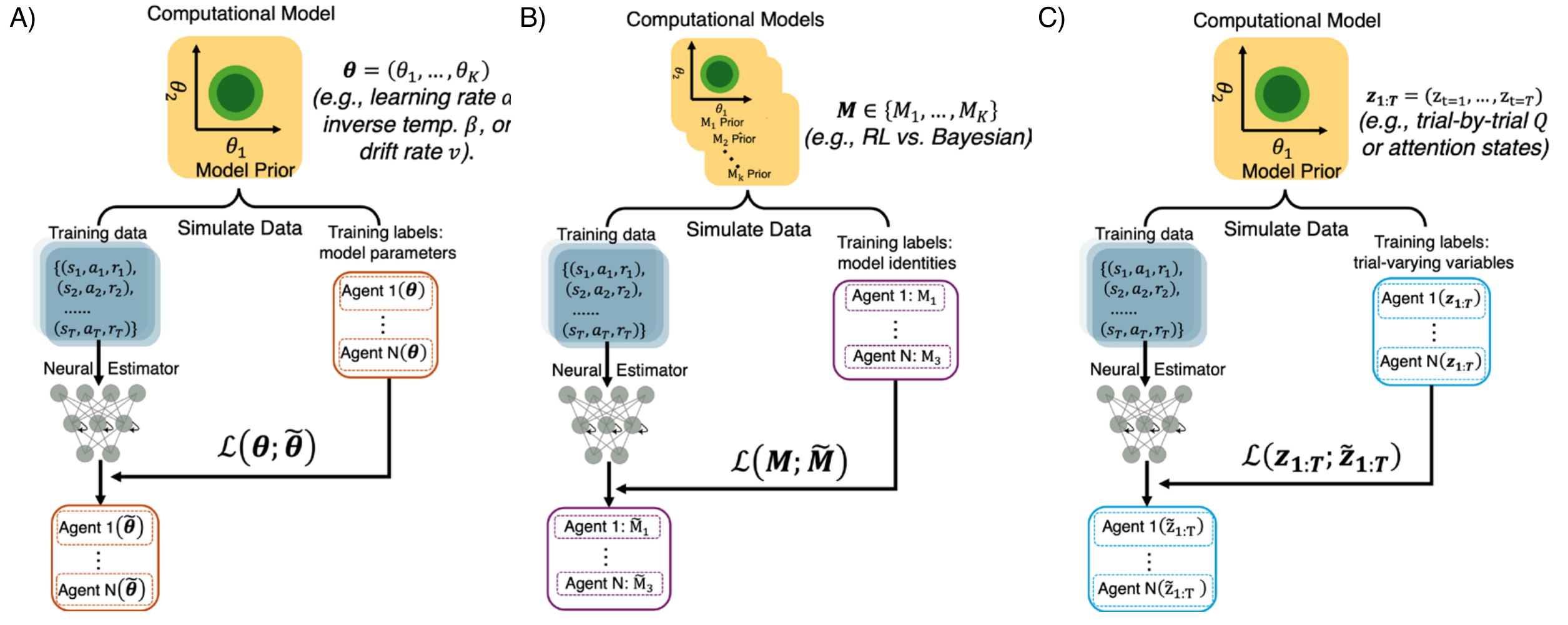} 
    \caption{\textbf{Use cases of neural SBI} A) Parameter estimation (\textbf{$\theta$}): Inferring static model parameters. B) Model comparison (\textbf{$M$}): Inferring which candidate model generated a given agent's data (e.g., comparing competing cognitive models $M_1, \dots, M_K$). C) Trial-varying latent variable inference (\textbf{$z_{1:T}$}): Inferring dynamic latent variables across trials.}    
    \label{fig:sbi_app}
    \vspace{-15pt}    
\end{figure}
Numerous studies have demonstrated the utility of neural SBI across diverse use cases, including parameter estimation, model comparison, and trial-varying latent variable inference (Fig.~\ref{fig:sbi_app}; \cite{radev2020bayesflow, rmus2024artificial, pan2025latent}). In this tutorial, we provide guidance across these three domains commonly used in cognitive modeling:
\begin{itemize}[leftmargin=2.5em, nosep]
    \item \textbf{Parameter estimation} identifies parameter values that best account for empirical data under a given model. These parameters can be used to computationally phenotype individual differences \parencite{katahira2021revisiting, xia2020learning, schurr2024dynamic} or to quantify the effects of experimental, pharmacological, or lesion-based interventions \parencite{frank2004carrot, lorains2014strategic, erfanian2024dopamine} on interpretable cognitive variables.
    \item \textbf{Model comparison} evaluates which candidate model among a set of alternatives best accounts for observed data, enabling formal adjudication between competing theoretical hypotheses or cognitive mechanisms \parencite{daw2011trial, collins2026habit, song2022humans}.
    \item \textbf{Trial-varying latent variable inference} captures trial-by-trial fluctuations in internal states to explain behavioral variability. For instance, state-space models such as hidden Markov models (HMM) can infer dynamic attentional states underlying choice behavior \parencite{ashwood2022mice, li2024dynamic}. In neuroscience, linking neural activity to model-derived latent variables illuminates how the brain supports cognition \parencite{cohen2017computational, eshel2016dopamine}, such as correlating reinforcement learning (RL) reward prediction errors (RPEs) with human ventral striatum fMRI signals \parencite{o2007model}.  
\end{itemize}

\subsection{When to consider neural SBI}
Despite the versatility of neural SBI for cognitive modeling, standard likelihood-based methods (e.g., MLE or MCMC) remain preferable when a model has a tractable likelihood, as neural SBI may not always offer superior accuracy or efficiency in such regimes \parencite{rmus2024artificial, kvam2025comparing}. We recommend adopting neural SBI primarily in two scenarios: \textbf{1) Intractable likelihoods}. Neural SBI circumvents the need to compute or approximate likelihoods. Unlike ABC, neural estimators remove the necessity of handcrafted summary statistics and arbitrary distance thresholding while consistently yielding superior performance \parencite{rmus2024artificial}. Thus, when the likelihood is intractable, neural SBI is generally a strong first choice of approach. \textbf{2) Amortized inference for large empirical datasets:} Amortization requires upfront training costs but enables near-instantaneous inference on subsequent datasets (e.g., many participants) without refitting the model for each individual. However, there is no simple rule of thumb for how much empirical data constitutes a ``large’’ dataset, as the benefits of amortization depend on simulation speed and the amount of synthetic data required. In prior work, simulating and training on 30,000 synthetic datasets using cognitive models with fewer than 20 parameters and data dimensions below 1,000 often takes on the order of minutes on an A10 GPU \parencite{deistler2025simulation, pan2025latent}. A practical guideline is therefore to consider neural SBI when the conventional inference approach requires several days to estimate the parameters of interest, particularly when the trained neural estimator will be reused across many datasets.
\section{Step-by-step guide}
\subsection{Experimental design and cognitive model development}
\begin{figure}[b!]
    \centering
    \includegraphics[width=1\linewidth]{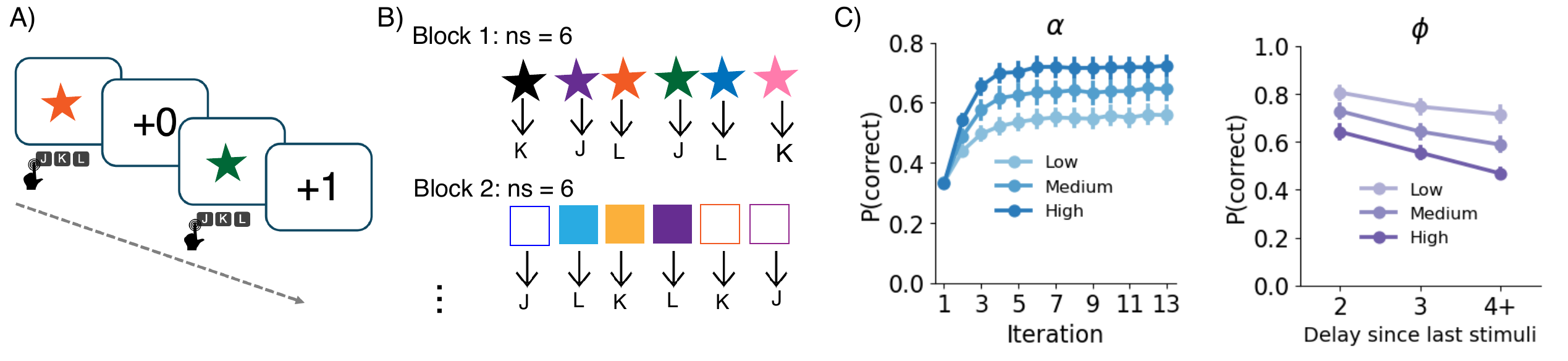}
    \vspace{-15pt}    
    \caption{\textbf{Illustrative example} A) A reward-based learning task. B) The associations between 6 stimuli and 3 actions are learned in each block. C) Simulation of the RLF model on the reward-based learning task, showing how different parameter values (learning rate $\alpha$ and forgetting rate $\phi$ impact different aspects of behavior: learning curves (left: accuracy as a function of stimulus iteration); forgetting (right: accuracy as a function of intervening trials).}    
    \label{fig:sbi_exp}
\end{figure}
Before applying neural simulation-based inference (SBI), researchers must ensure that both the experimental task and the candidate cognitive models align with the target scientific questions \parencite{wilson2019ten}. Here, we use a reward-based learning task to demonstrate how relevant experimental features are extracted to train neural estimators, though the principle readily generalizes to other paradigms. We chose this task and model to balance simplicity with sufficient complexity to highlight common neural SBI challenges and solutions. These include multi-block structures, multi-trial dependencies, and partially overlapping parameter effects on behavior. Notably, selecting a likelihood-tractable model allows us to directly benchmark neural SBI against widely used techniques.

In our example task, agents (e.g., participants) learn stimulus-action associations via trial and error (Fig.~\ref{fig:sbi_exp}A). On each trial, the agent responds to a stimulus (an image) by choosing one of three keys (J, K, L), receiving deterministic binary feedback ($r_t \in \{0, 1\}$), indicating correctness.  The goal of the agent is to earn as many points as possible. The full task comprises 12 independent blocks; within each block, the agent learns mappings across 6 distinct, novel stimuli and 3 actions, with each stimulus presented for 13 iterations, pseudo-randomly interleaved (Fig.~\ref{fig:sbi_exp}B; see \textcite{collins2026habit} for examples of similar experimental design in empirical data).

We use a standard reinforcement learning (RL) framework to model this reward-based learning task \parencite{niv2012neural, eshel2016dopamine,sutton2018reinforcement}. Specifically, we assume that agents track stimulus-action associations $Q(s,a)$. After observing stimuli, actions, and rewards $(s_t, a_t, r_t)$ at trial $t$, Q-values are updated through a classic delta rule:
\noindent{}
\begin{equation}
\begin{gathered}
\delta_t = r_t - Q_t(s_t, a_t) \\
Q_{t+1}(s_t, a_t) = Q_t(s_t, a_t) + \alpha*\delta_t
\end{gathered}
\end{equation}
\noindent{}
Values are initialized uniformly to $Q_0 = \frac{1}{n_A}$, where $n_A$ denotes the total number of possible actions in a task (three in the illustrated example), and $\alpha \in [0, 1]$ denotes a learning rate. In addition, we incorporate counterfactual updating in this model, where the values of the non-chosen actions also get updated on each trial \parencite{eckstein2022reinforcement}:
\noindent{}
\begin{equation}
\begin{gathered}
\delta_\text{unchosen} = (1-r) - Q_{t}(1-a) \\
Q_{t+1}(1-a) = Q_{t}(1-a) + \alpha*\delta_\text{unchosen}
\end{gathered}
\end{equation}
\noindent{}

Action selection is governed by a softmax policy with inverse temperature $\beta \in [4, 16]$. A``sticky'' parameter $\kappa \in [-1, 1]$ is used to account for stimulus-independent choice perseveration -- the tendency to repeat the preceding response regardless of the presented stimulus:
\noindent{}
\begin{equation}
\begin{gathered}
\pi_{\text{RL}}(a \mid s) = \frac{\exp\left(\beta Q_t(s, a) + \kappa I_t(a, a_{t-1})\right)}{\sum_i \exp\left(\beta Q_t(s, a_i) + \kappa I_t(a, a_{t-1})\right)}, \\
\text{where } I_t (a,a_{t-1}) = 
\begin{cases} 
1, & \text{if } a = a_{t-1}, \\[-6pt]
0, & \text{if } a \neq a_{t-1}.
\end{cases}
\end{gathered}
\end{equation}
\noindent{}

Finally, the model updates stimulus-action association strengths immediately after experiencing outcomes but is susceptible to memory decay. All stimulus-action values decay toward their baseline initialization at the end of each trial:
\noindent{}
\begin{equation}
\begin{gathered}
\forall (s, a), Q_{t+1}(s, a) = Q_t(s, a) + \phi \left(Q_0 - Q_t(s, a)\right)
\end{gathered}
\end{equation}
\noindent{}
where $\phi \in [0, 1]$ denotes a forget rate parameter. This reinforcement learning with forgetting (RLF) thus includes four free model parameters: learning rate ($\alpha$), forget rate ($\phi$), inverse temperature ($\beta$), and choice stickiness ($\kappa$). We chose RLF, a model with a tractable likelihood, to provide a likelihood-based benchmark (i.e., MLE/MAP) against neural SBI in this tutorial.

\subsection{Generate synthetic data}
Once a well-defined experimental design and candidate cognitive models are established, training a neural estimator requires generating synthetic data under the target paradigm; thus, all candidate models must be forward-simulable. Parameter priors must also be specified with care: because neural estimators learn the statistical distribution of the training data, the prior is implicitly embedded into the inference process. Consequently, misspecified priors can substantially degrade neural SBI performance \parencite{gao2023generalized}. Unless strong domain knowledge justifies informative priors, we recommend using broad priors that cover the full space of plausible parameter values, while still enforcing theoretical and physiological constraints so that simulated outputs remain consistent with realistic participant behavior, and the model operates in a theoretically meaningful range.

With the specified generative model and prior distributions, the next practical consideration is the simulation budget. Neural networks are highly parameterized and require large simulated data sets to train. However, it is best practice to begin with a modest training dataset (e.g., $1\text{k}$--$3\text{k}$ simulated agents per model) to verify the pipeline integrity and baseline parameter recovery, alongside a dedicated test set for evaluation (e.g., $5\text{k}$ agents per model). The training budget can later be scaled up to optimize estimation accuracy. Unlike parameter estimation and trial-varying latent variable inference, model comparison requires generating synthetic data across all candidate models. 

\subsection{Validate Synthetic Data}
Before training neural networks, the synthetic data should undergo a sanity check using predefined qualitative behavioral metrics. Specifically, simulated behavioral patterns should overlap with observed or expected empirical observations to ensure that the prior is broad enough to capture meaningful behavioral variance ~\parencite{wilson2019ten}. In practice, this entails evaluating the simulated data along informative summary statistics. When crafting these metrics, it is particularly helpful to identify measures that isolate the behavioral footprints of individual parameters, as these directly inform downstream feature selection. For example, in the RLF model, examining choice accuracy across stimulus repetition (learning curves) and as a function of the delay since a stimulus was last encountered (Fig.~\ref{fig:sbi_exp}C) reveals the distinct impact of each parameter on behavior. As expected, higher $\alpha$ and lower $\phi$ values systematically yielded higher accuracy, confirming that the simulations captured the target behavioral dynamics.

\subsection{Extract Features and Labels}
\begin{figure}[t!]
    \centering
    \includegraphics[width=1\linewidth]{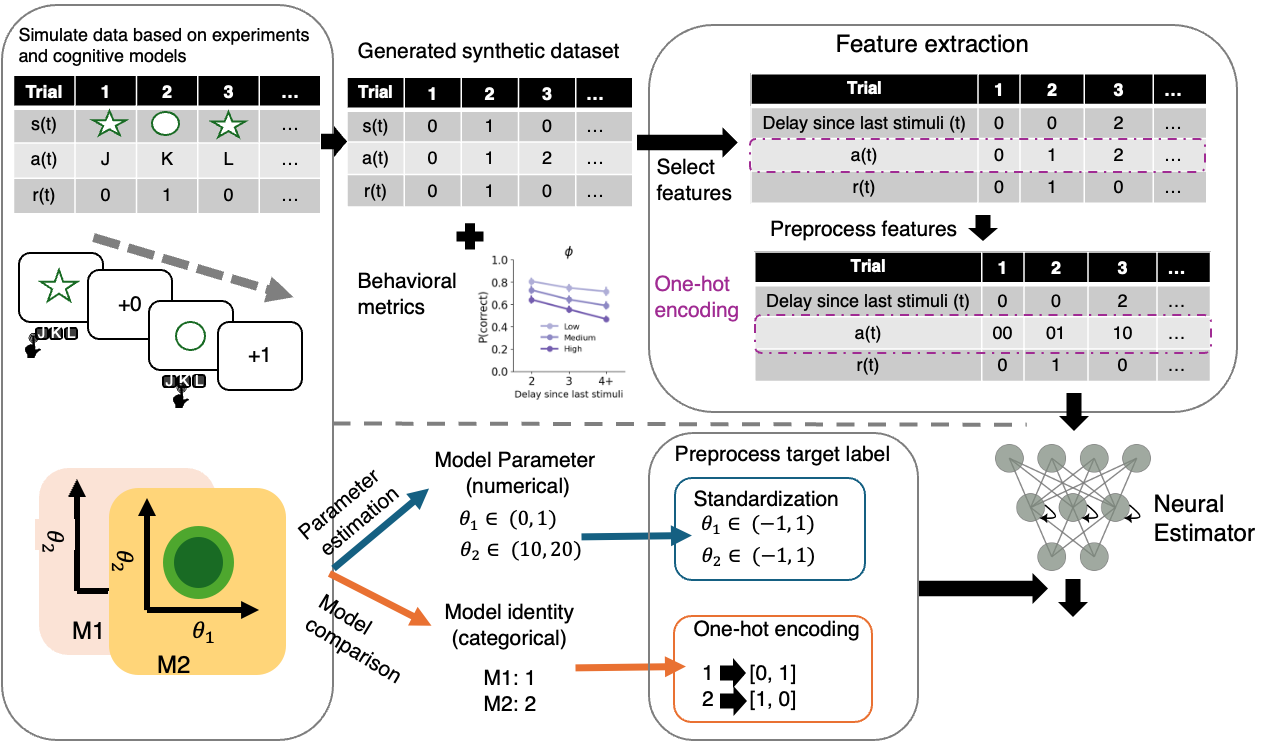}  
    \caption{\textbf{Input feature and target label extraction.} \textbf{Top row:} Input features are selected based on the task structure and cognitive models, then preprocessed accordingly (e.g., one-hot encoding nominal action features). \textbf{Bottom row:} Target labels are defined by the inference goal, whether estimating numerical model parameters (blue arrow) or performing model comparison (orange arrow). These labels are then preprocessed using standard techniques, such as standardizing numerical parameters or one-hot encoding nominal model identities.}    
    \label{fig:features_process}
    \vspace{-20pt}      
\end{figure}

Once the synthetic data are validated, they should be formatted into input features and target labels suitable for ANN training. Input features represent the information fed into the network; these must consist exclusively of observable variables that researchers can collect in empirical experiments (e.g., stimuli, choices, outcomes, and response times). Conversely, target labels are the outputs the ANN is trained to infer. For synthetic data, these target labels are known; in real data, they represent the unobservable quantities, such as latent variables, model parameters, or model identities, that researchers aim to infer from real participants. During training, the ANN learns to map input features to labels using the known, ground-truth labels from the synthetic data. At inference time, the trained network takes real participant data, where only the observable input features are available, to infer the corresponding unknown parameters or latent variables. Below, we outline best practices for selecting and preprocessing both features and labels (Fig.~\ref{fig:features_process}; \cite{kotsiantis2006data}).

\paragraph{Feature selection} Feature selection critically governs training efficiency. Though in theory, ANNs can distill necessary representations if all experimental information is present, in practice, a poor input representation can substantially degrade performance. Because neural SBI can be simulation-intensive, supplying informative features, even if potentially redundant, can reduce the required simulation budget and lower the computational overhead of both simulation and training. Feature selection is guided by two primary considerations:
\begin{itemize}[leftmargin=2.5em, nosep]
    \item \textbf{The relationship between cognitive model and data} Input features that provide highly diagnostic information to differentiate parameter values are useful. For instance, in the RLF model (Fig.~\ref{fig:sbi_exp}C), identifying the effect of the learning rate $\alpha$ requires $\texttt{reward}$ and $\texttt{stimulus\_iteration}$, whereas the forget rate $\phi$ impacts performance as a function of $\texttt{delay\_since\_last\_stimulus}$, making this input feature highly informative. 
    \item \textbf{Experimental structure:} Features should include a minimal, parsimonious set of variables encompassing task conditions/manipulations and block structures. Multi-condition designs (e.g., win/loss, go/no-go) should be represented as categorical features. In block-based designs, a binary indicator marking block transitions (e.g., $\texttt{is\_switch}$) adequately captures task structure without requiring explicit block or trial indices, as sequential architectures (such as recurrent neural networks) naturally track temporal progression.
\end{itemize}
\paragraph{Feature pre-processing} Following feature selection, input features must be preprocessed to ensure stable neural network optimization (Fig.~\ref{fig:features_process} top row). Continuous input features (e.g., reward value, stimulus discriminability) should generally be scaled to comparable ranges, typically via standardization (zero mean, unit variance) or min–max normalization, to ensure reliable convergence and prevent large-magnitude features from disproportionately driving gradient updates \parencite{lecun2002efficient, sola1997importance}. Categorical variables require numerical encoding: ordinal features with meaningful rankings are integer-encoded (e.g., block number), whereas nominal features are one-hot encoded (\cite{hancock2020survey}; See Appendix Table~\ref{app:feature} for more details). 

\paragraph{Target labels selection and pre-processing} For parameter estimation or trial-varying latent variables inference, the target labels represent the unobservable values that researchers aim to infer from empirical data (Fig.~\ref{fig:features_process} bottom row). Depending on whether the targets are continuous or discrete, they undergo the same scaling or encoding steps applied to the input features. For model comparison, each model identity acts as a discrete class and is thereby one-hot encoded as a nominal target.

Lastly, as is standard practice in machine learning, neural estimators learn by iteratively updating their weights to minimize an \emph{objective function} (a measure of prediction error; see below), via mini-batch gradient descent \parencite{lecun2002efficient}. Rather than processing the entire dataset at once, this approach updates the network's weights incrementally on small subsets of data called \emph{mini-batches}. Because simulations are often generated sequentially (e.g., simulating all agents from Model 1 before Model 2, or sweeping systematically through parameter space), leaving the dataset in its original order would result in mini-batches that are homogeneous. When exposed to such unvaried batches, the network can over-adapt to local data patterns, causing optimization to stall in suboptimal solutions (i.e., local minima).  To prevent this, the dataset must be randomly shuffled across simulated agents prior to training. Importantly, shuffling occurs at the agent level, keeping each agent's behavioral trajectory paired with its corresponding ground-truth label ($M_i, \theta_i$), rather than across trials within an agent, thereby preserving the temporal order of trials.

\subsection{Train neural estimators}
With the data prepared, training a neural estimator requires specifying two core components: the network architecture and the objective (loss) function. These choices should be guided by the structure of the input data and the type of inferential output researchers aim to obtain.

\paragraph{Network structures}
Neural estimators typically comprise two distinct sub-networks: an \emph{embedding network} that compresses input data into an informative summary embedding, and an \emph{inference network} that maps this summary embedding to the target outputs. The choice of embedding network depends on the structure of the input data. For example, convolutional neural networks (CNNs) are well-suited for spatial or image data \parencite{lenzi2023neural}, as well as trial-independent data (e.g., decision-making task data \cite{radev2020towards, fengler2021likelihood}). Conversely, recurrent neural networks (RNNs) capture sequential dependencies for time-series data \parencite{ger2023harnessing}. Finally, for input data lacking a clear structure, multilayer perceptrons (MLPs) serve as a good starting point; as simple feedforward neural networks, they learn nonlinear transformations of high-dimensional inputs to produce compact representations for downstream inference.

By contrast, the choice of inference network typically depends on the complexity of the target outputs. Because most cognitive models involve only a small number of parameters, standard MLPs are commonly used \parencite{kvam2025comparing, rmus2024artificial}. For high-dimensional target spaces, such as estimating full joint posterior distributions or brain imaging models, researchers can use more advanced inference structures like normalizing flow \parencite{hashemi2024simulation, radev2020bayesflow}. In our tutorial example, we paired an RNN embedding network with an MLP inference network to estimate parameters of the RLF model, following established architectures for sequential behavioral data \parencite{dezfouli2019disentangled, rmus2024artificial, pan2025latent}. To make these methods accessible, packages like \href{https://bayesflow.org/v2.1.0/index.html}{BayesFlow} provide APIs and examples, allowing researchers to easily configure these networks without coding them from scratch.

\paragraph{Loss functions}
Complementing the network architecture, the objective/loss function, which measures how well the network prediction matches the target outcome, must align with both the mathematical format of the target variable and the desired inferential goal. For discrete or categorical targets (e.g., model identity or trial-by-trial attention states), networks are often optimized using a fit measure called \emph{categorical cross-entropy}. For continuous targets (e.g., learning rates or trial-by-trial RPEs), \emph{mean squared error} (MSE) is standard for point estimation, whereas \emph{Kullback–Leibler (KL) divergence} can be used to approximate the full posterior distribution \parencite{mittal2025context}. In practice, packages like \href{https://bayesflow.org/v2.1.0/index.html}{BayesFlow} automatically select an appropriate default loss function based on the chosen network architecture and target variables, while still allowing researchers to customize it as needed.

\paragraph{Training and validation}
\begin{wrapfigure}[7]{r}{0.3\columnwidth}
    \vspace{-10pt}
    \centering
    \includegraphics[width=0.3\columnwidth]{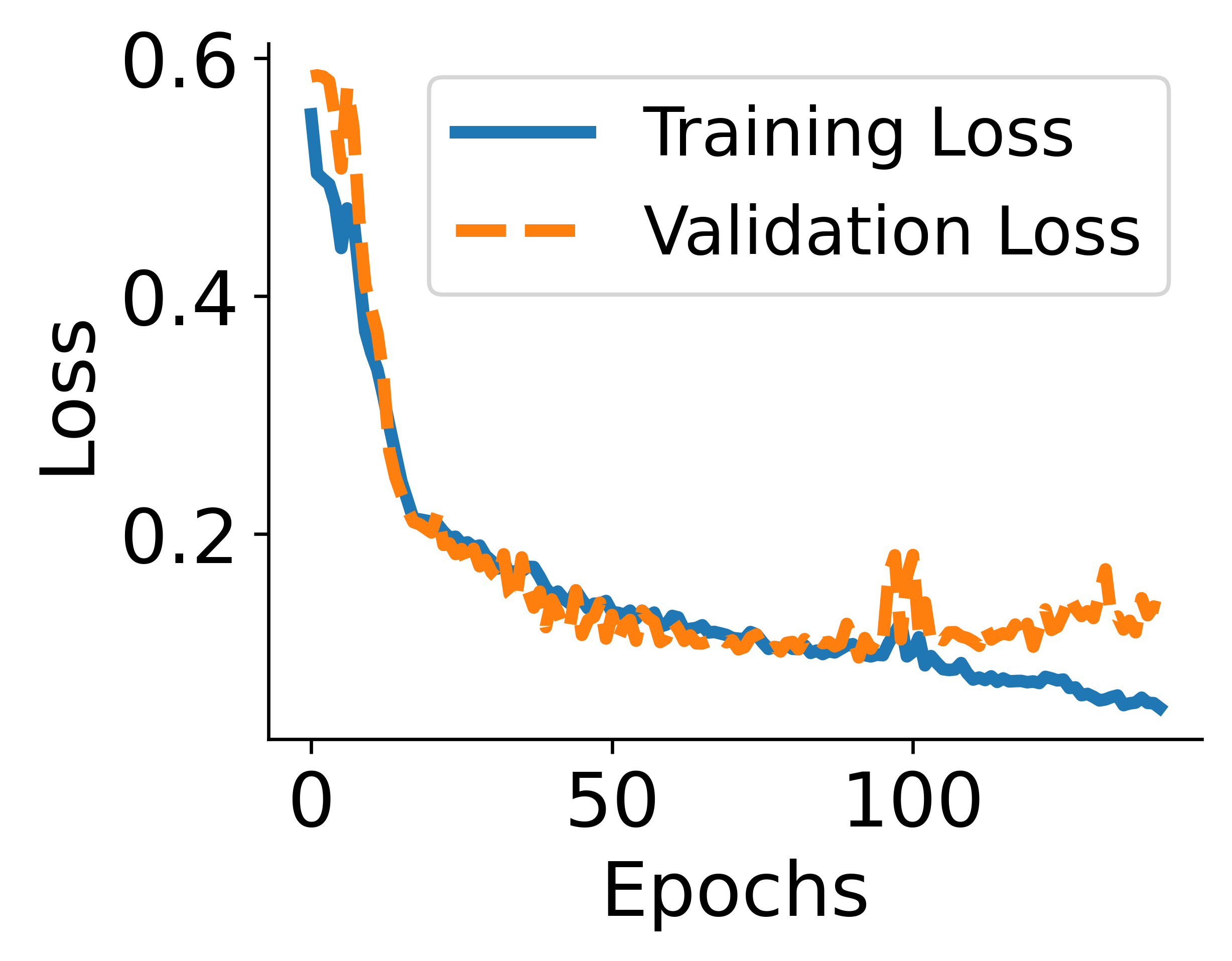} 
    \caption{\textbf{Training and validation loss example}} 
    \label{fig:loss_plot}
    \vspace{-15pt}
\end{wrapfigure}

After the network architecture and objective function are specified, the preprocessed dataset should be partitioned into training and validation (in addition to a separate held-out test set) prior to fitting the neural estimator. During training, the training set is used to iteratively update the network weights, while the validation set provides an independent check against overfitting. Ideally, both training and validation losses should decrease together (Fig.~\ref{fig:loss_plot}); if the validation loss begins to plateau or rise while the training loss continues to fall, the network is starting to overfit. To prevent this, techniques such as early stopping, which halts training once validation performance stops improving, are strongly recommended (\href{https://github.com/ti55987/DL4CM/blob/master/parameter_estimation_workflow.ipynb}{code example}). Much like configuring network architectures and loss functions, these safeguards are built directly into mature toolkits (e.g., \href{https://bayesflow.org/v2.1.0/index.html}{BayesFlow}, \href{https://sbi-dev.github.io/sbi/v0.25.0/}{sbi}), making neural SBI accessible without requiring researchers to implement them from scratch.

\subsection{Evaluate neural estimators}
\begin{figure}[t!]
    \centering
    \includegraphics[width=1\linewidth]{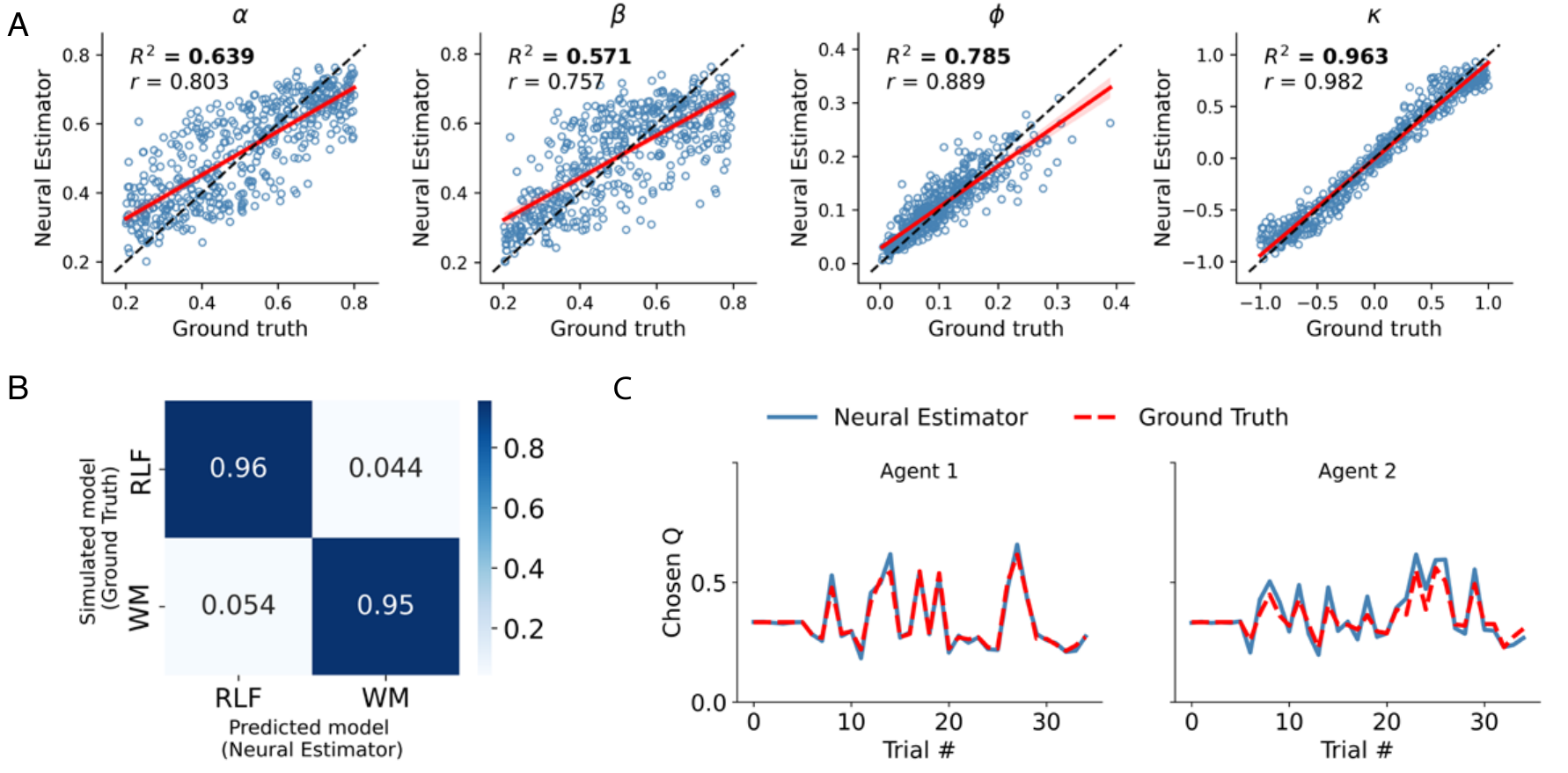} 
    \caption{\textbf{Example qualitative measurements across three use cases.} A) Parameter correlation between ground truth parameter and inferred parameter shows good recoverability. B) Confusion matrix between an RLF and a working memory model (WM) shows good model identifiability. C) Trial-by-trial chosen Q values from an RLF model, for the ground truth of simulated data (red) and inferred from simulated data (blue). The neural estimator is based on the LaseNet \cite{pan2025latent} Experiment details can be found in Appendix~\ref{sup:exp_details}.}    
    \label{fig:qualitative_res}
    \vspace{-25pt}    
\end{figure}
\begin{figure}[b!]
    \centering
    \includegraphics[width=1\linewidth]{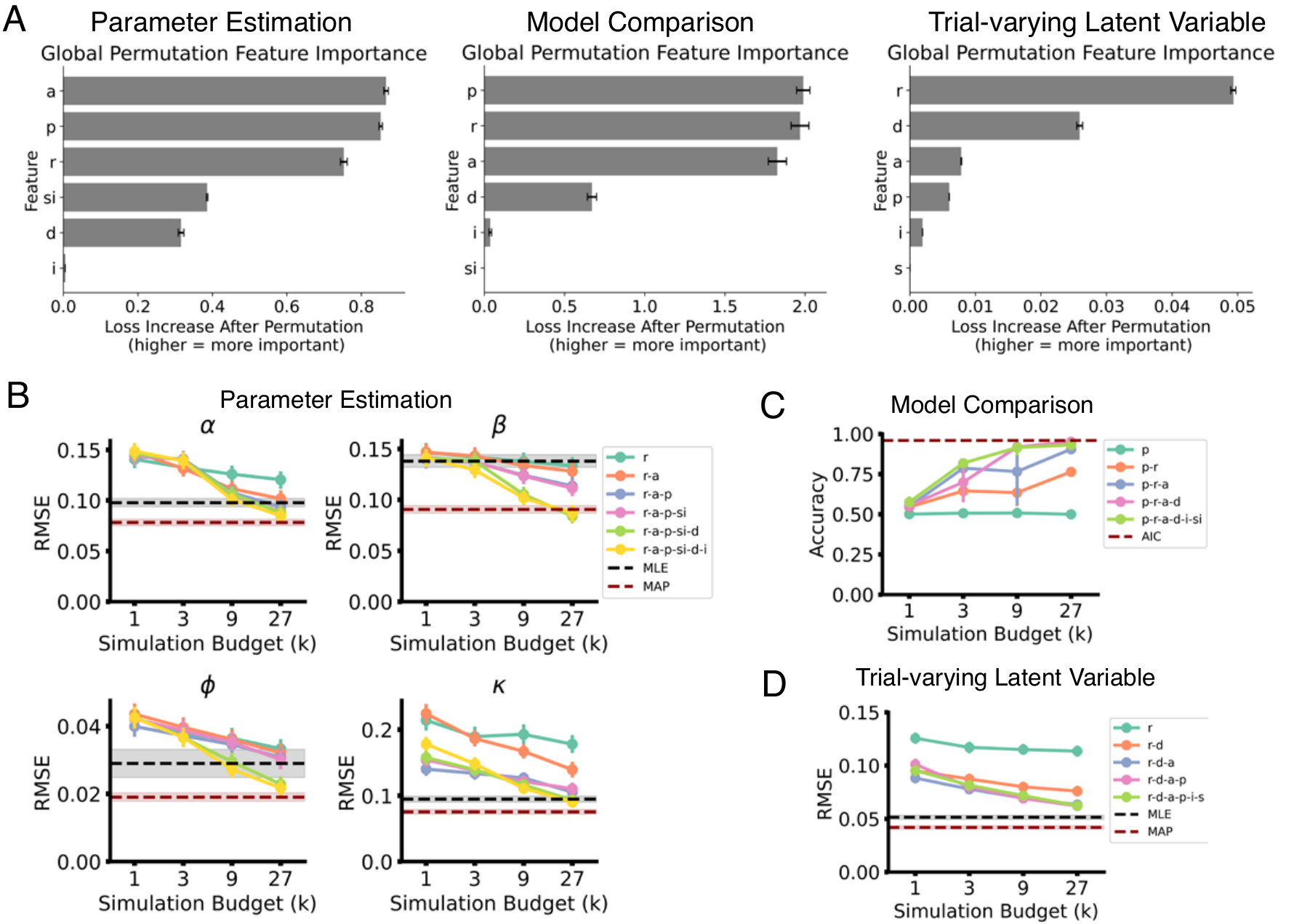}
    \vspace{-15pt}    
    \caption{\textbf{A)} Feature importance ranking. Left: parameter estimation on an RLF model. Middle: model comparison between RLF and WM models. Right: trial-varying RPE inference. \textbf{B-D)} Quantitative results of three use cases across different simulation budgets and combinations of features. \textbf{B)} RMSE of four different parameters in an RLF model. \textbf{C)} Accuracy of model comparison between RLF and WM models. \textbf{D)} RMSE of trial-varying RPE inference. a: action, p: previous action, r: reward, si: stimulus iteration, s: stimulus, d: delay since last stimulus, i: is switch. Experiment details can be found in Appendix~\ref{sup:exp_details}.}    
    \label{fig:sbi_parameter}
\end{figure}
Once trained, neural estimators must be systematically evaluated on the held-out test dataset using both quantitative and qualitative assessments. Quantitatively, metrics like root mean squared error (RMSE) or mean absolute error (MAE) are standard for continuous variables, whereas negative log loss (NLL) or classification accuracy are suitable for discrete variables. Qualitatively, comparing predictions against the ground truth using correlation plots or confusion matrices for model comparison can help identify where the neural estimator fails (e.g., which ranges of parameter values are difficult to recover and why; Fig.~\ref{fig:qualitative_res}). If recovery is poor, several diagnostic and optimization strategies can be employed. Here, we outline a few simple and commonly used approaches. 

\paragraph{Permutation tests for feature importance} Often, researchers do not know in advance which features are most informative for training a neural estimator. We therefore recommend starting with a broad but plausible set of features, and applying post-hoc feature-importance analyses to assess their relative contributions. The results can then be used to iteratively refine the feature set (returning to Step 3 of the workflow). A canonical method is the permutation feature-importance test, which proceeds in five steps:  (1) evaluate the trained neural estimator on an held-out test dataset to establish baseline performance; (2) randomly permute one input feature while leaving all other features intact in test dataset; (3) evaluate the neural estimator on the permuted dataset; (4) quantify the resulting drop in performance relative to the baseline; and (5) repeat step 2-4 across all features to establish global feature importance (\cite{breiman2001random}; \href{https://github.com/ti55987/DL4CM/blob/master/parameter_estimation_workflow.ipynb}{code example}).

Often, features dictated strictly by experimental design, such as stimuli, tend to be less informative (i.e., lower global importance) than features whose values or temporal dynamics depend on the cognitive model and its underlying parameters. In the RLF model, `stimulus iteration`, `delay since last stimulus`, and `is switch` show lower feature importance because their combinatorial space is constrained by the static task structure (typically less than the order of $10^3$ unique sequences; Fig.~\ref{fig:sbi_parameter}A). Conversely, features such as `action` and `reward` depend on the cognitive model or its underlying parameters, thus exhibiting much more diverse sequences, easily exceeding the order of $10^4$ distinct patterns, especially when parameters are continuous or high-dimensional. Consequently, though `reward` and `stimulus iteration` are critical to recover the learning rate parameter, the reward sequence is more informative because its model-dependent variability provides a richer signal for distinguishing between different learning rates.

To illustrate how different features contribute to parameter recovery in the RLF model, we tested several feature combinations based on their permutation-importance rankings across varying simulation budgets. Consistent with its low importance ranking, adding `is switch` yielded negligible performance gains across all parameters, likely because the network can infer block transitions from `stimulus iteration`. Feature ablation analyses confirmed that omitting `is switch` had no detectable impact on overall recovery, whereas replacing `stimulus iteration` with raw `stimuli` markedly degraded recovery of the learning rate $\alpha$ (Appendix Table~\ref{tab:feature_ablation}). By contrast, model-dependent features selectively inform certain parameters: adding `delay since last stimulus` substantially improved the recovery of the forget rate $\phi$  (RMSE: 0.03 for `r-a-p-si` pink line vs. 0.022 for green line at 27k in Fig.~\ref{fig:sbi_parameter}B).  Similarly, `previous action` selectively enhanced recovery of choice stickiness (RMSE: 0.139 for `r-a` orange line vs. 0.105 for `r-a-p` blue line at 27k in Fig.~\ref{fig:sbi_parameter}B; confirmed by ablation results in Appendix Table~\ref{tab:feature_ablation}). 

Importantly, identifying an optimal feature set can reduce the simulation budget required to achieve target performance. In the model comparison between RLF and WM models, a four-feature set of `previous action`, `rewards`, `actions`, and `delay since last stimulus` reached high classification accuracy at one-third of the simulation budget required by a three-feature set that omitted `delay since last stimulus` (Accuracy: 91.86\% for `p-r-a-d` pink line at 9k vs. 90.5\% for `p-r-a` blue line at 27k in Fig~\ref{fig:sbi_parameter}C). Notably, having more features doesn't mean better performance: the feature set with and without `is switch` and `stimulus iteration` converged by 9k simulation budget and reached nearly ceiling accuracy at 27k (Accuracy: 95.1\% for `p-r-a-d` pink vs. Accuracy: 93.2\% for `p-r-a-d-i-si` light green lines at 27k. Accuracy of AIC is 95.9\% in Fig~\ref{fig:sbi_parameter}C). Similarly, when inferring trial-varying RPEs in the RLF model, three features `rewards`, `delay since last stimulus`, and `actions` are sufficient to reach similar performance to the six-feature set (RMSE:0.063 for `r-d-a` blue line vs. 0.062 for `r-d-a-p-i-s` light green line at the 27k simulation budget in Fig~\ref{fig:sbi_parameter}D).

\paragraph{Hyperparameter tuning} Hyperparameters are settings that determine how a neural network is structured and trained, such as batch size, learning rate, or network units. Tuning these hyperparameters can help the network find a good solution more efficiently for a given dataset. Common approaches include random search (starting from reasonable parameters that work for similar problems) or Bayesian optimization to algorithmically search for effective hyperparameter settings \parencite{snoek2012practical, bergstra2013making}. Several software packages can automate this process, such as \href{https://keras.io/keras_tuner/}{keras-tuner} (\href{https://github.com/ti55987/DL4CM/blob/master/parameter_estimation_workflow.ipynb}{example code}; \cite{omalley2019kerastuner}) or \href{https://optuna.org./}{optuna}\parencite{optuna_2019}.

\paragraph{Increasing training dataset} Expanding the simulation budget exposes the estimator to a broader distribution of plausible data patterns. Once the training pipeline is validated, the training dataset size can be scaled up incrementally. As shown in Fig~\ref{fig:sbi_parameter}B, increasing the dataset to 9k and 27k agents substantially improves parameter recovery, bringing the RMSE to levels comparable with standard likelihood-based MAP estimation. Nonetheless, there is no universal rule for how much simulation is required, as the optimal simulation budget depends on the complexity of both the experiment and the cognitive model \parencite{deistler2025simulation}. Prior applications in cognitive modeling indicate that $\sim$30,000 simulated agents are generally sufficient to achieve good performance, with few exceptions that may require an order of 100k simulations \parencite{rmus2024artificial, pan2025latent, kvam2025comparing}. We also recommend saving the simulated dataset so that it can be reused for subsequent iterations or other use cases. If the dataset becomes so large that loading it from the storage slows down training, several strategies can improve data-loading efficiency, including chunking the dataset for sequential streaming or adopting efficient binary/columnar storage formats such as HDF5 (\href{https://docs.h5py.org/en/latest/quick.html}{example usage}) rather than CSV/JSON.

Lastly, to safeguard against convergence to suboptimal local extrema, neural estimators should be trained and evaluated across multiple random initializations. Once satisfactory recovery is confirmed on a synthetic test dataset, the neural estimator can be applied to infer variables of interest from real participant data.

\subsection{Fit empirical data and report}
To fit empirical data, researchers should extract the same input features defined in workflow Step 3 for neural estimators. Depending on the target use case, the neural estimators then infer parameters, model identities, or trial-by-trial latent variables. Because inference is amortized, this process requires only milliseconds. In line with general computational modeling best practices, the inferred parameters, model identities, and trial-varying latent variables, should not be simply accepted as true. Instead, researchers should conduct further validation, including posterior predictive checks, to ensure the obtained results are interpretable and meet absolute thresholds for the researcher's goals, including qualitatively capturing phenomena of interest ~\parencite{wilson2019ten, palminteri2017importance}. For neural posterior estimation, several diagnostic techniques can additionally assess posterior quality in the absence of ground-truth values \parencite{lueckmann2021benchmarking, deistler2025simulation}.

Finally, researchers should report the recovery results obtained from Step 5 \textit{Evaluate neural estimator}. To ensure reproducibility, include comprehensive training details, such as hyperparameter configurations, code, and computing environments (Appendix~\ref{sup:exp_details}). 

\section{Discussion}
Neural SBI provides a powerful framework for overcoming intractable likelihoods in computational models \parencite{cranmer2020frontier, gloeckler2024all}. Although several studies have highlighted the utility of neural SBI to expand the space of testable cognitive models \parencite{schumacher2025validation, maher2026learning, leng2026perils}, its broader adoption remains limited by a lack of accessible, domain-specific guidance. To bridge this gap, we presented a practical tutorial on neural SBI tailored to cognitive modelers. Using simple RL-based models as accessible testbeds, we illustrated how to select features to mitigate computational and simulation bottlenecks (see Appendix~\ref{sup:intractable_example} for an example model with an intractable likelihood).
 
In addition to expanding the hypothesis space with likelihood-intractable models, neural SBI offers fast inference speed, making it highly scalable to large empirical datasets. Furthermore, prior studies have suggested that neural SBI approaches outperform standard information criteria (e.g., AIC and BIC) in model comparison, even when likelihoods are tractable \parencite{rmus2024artificial, kvam2025comparing}. Nonetheless, more extensive testing is required to systematically characterize the boundary of this advantage across complex experimental designs, such as varying stimulus sequences or multiple task conditions.

\paragraph{Other practical considerations} In practice, we strongly recommend a start-small workflow. Before scaling up to large simulation budgets, practitioners should validate the end-to-end pipeline on small synthetic datasets (e.g., 3,000 to 6,000 simulated agents). This allows for rapid debugging and iterative refinement of feature sets with minimal computational overhead. Practitioners new to neural SBI can use our accompanying \href{https://github.com/ti55987/DL4CM}{example code} as an accessible starting point.

In addition, mature open-source software packages have substantially lowered the barrier to entry. Toolkits such as \texttt{BayesFlow} \parencite{kuhmichel2026bayesflow} and \texttt{sbi} \parencite{tejero-cantero2020sbi, BoeltsDeistler_sbi_2025} offer a flexible, accessible application programming interface (API) to implement diverse neural SBIs. Notably, \texttt{BayesFlow} supports multiple deep learning backends (PyTorch, TensorFlow, and JAX) along with curated cognitive modeling examples. Regarding hardware, utilizing a Graphics Processing Unit (GPU) is critical for efficient network training; cloud-hosted environments like Google Colaboratory (Google Colab) provide an accessible platform with pre-installed deep learning libraries and configurable GPU options.

While this tutorial primarily focused on neural posterior estimation (NPE) given its popularity in cognitive modeling, neural likelihood estimation (NLE) stands as a compelling alternative. Rather than targeting the posterior directly, NLE trains a neural estimator to approximate the likelihood function \parencite{fengler2021likelihood}. Practitioners can subsequently couple this approximated likelihood with standard posterior inference methods, such as MCMC or variational inference (VI) \parencite{zhang2018advances}. A key advantage of NLE is its prior flexibility: different prior distributions can be evaluated post-training without retraining the network. However, NLE introduces challenges in computing surrogate likelihoods for supervised learning, while the computational cost of downstream posterior sampling remains \parencite{Fengler_2026}.

\paragraph{Limitation and future work} Despite the swift maturation of the SBI ecosystem and its supporting software packages \parencite{kuhmichel2026bayesflow, tejero-cantero2020sbi}, several limitations stand out: First, applying neural SBI to high-dimensional data (e.g., complex stimulus spaces like a large chessboard) or highly parameterized models (e.g., exceeding 100 parameters) remains computationally demanding. Recent studies have proposed transformer-based architectures to handle these scenarios \parencite{gloeckler2024all}, but such networks often require substantial computational resources. Furthermore, generating sufficient synthetic training data becomes a bottleneck when the simulator itself is computationally expensive. While this tutorial highlighted feature selection as a practical mitigation, future work should explore automated and iterative feature selection tailored to domain-specific psychometrics.

Second, neural SBI pipelines remain largely specific to a single cognitive model and experimental design, requiring retraining whenever the cognitive model or task structure changes (e.g., varying stimulus counts or task manipulations). While recent frameworks like CogFormer \parencite{huang2026cogformer} demonstrate the feasibility of training a single neural estimator across multiple nested or related model variants (e.g., within the drift diffusion model family), generalizing across heterogeneous model classes (e.g., drift diffusion model and RL model classes) and distinct task structures remains an open challenge. Extending the neural SBI tool to train once for diverse models or experiments will greatly facilitate rapid cognitive model iteration.

Finally, the current workflow typically necessitates training distinct neural estimators for different inference goals: parameter estimation, model comparison, and trial-by-trial latent variable inference. However, several alternatives can reduce this overhead when a research question requires multiple inferences. For example, NLE allows the same approximated likelihood trained for parameter estimation to be reused for standard information-criterion model comparison (e.g., AIC/BIC) without training an additional neural estimator. Furthermore, emerging neural SBI methods have begun exploring joint posterior estimation to recover both static model parameters and dynamic latent variables simultaneously \parencite{aushev2024likelihood, ghosh2024sample}. Similarly, when trial-varying latent states (e.g., hidden Markov models) render a model's likelihood intractable, neural estimators such as LaseNet \parencite{pan2025latent} can infer latent state trajectories directly; this restores a tractable marginal likelihood and enables conventional likelihood-based methods to handle parameter estimation without training a separate estimator.

\paragraph{Further reading and advanced topics} While this tutorial covers the core foundations of neural SBI, practitioners ready to deepen their expertise can explore several advanced methodologies in the broader literature. 
\begin{itemize}[leftmargin=2.5em, nosep]
    \item \textbf{Prior and model misspecification:} Methods for detecting out-of-distribution empirical data and ensuring robust inference when the true data-generating process deviates from the assumed priors or cognitive model \parencite{ward2022robust, schmitt2023detecting, gao2023generalized}.
    \item \textbf{Uncertainty quantification and posterior diagnostics:} Systematic benchmarking frameworks, such as simulation-based calibration (SBC), classifier two-sample tests (C2ST), and posterior predictive checks, to validate full posterior distributions rather than point estimates alone \parencite{lueckmann2021benchmarking, deistler2025simulation}.
    \item \textbf{Alternative SBI paradigms and architectures:} \textit{neural likelihood estimation (NLE)} \parencite{fengler2021likelihood}, neural ratio estimation (NRE) \parencite{hermans2020likelihood}. For handling higher-dimensional data,  practitioners can look further into \textit{diffusion-} \parencite{gloeckler2024all} or \textit{flow-matching-based SBI} \parencite{wildberger2023flow}.
    \item \textbf{General machine learning best practices:} Practical guidelines on neural network architecture design, regularization, and hyperparameter optimization \parencite{goodfellow2016deep, lecun2002efficient}.
\end{itemize}

\section{Conclusion}
This tutorial provides a step-by-step workflow, concrete use cases, and practical strategies to reduce computational and simulation overhead in applying neural SBI. Ultimately, this guide aims to lower the barrier to entry, empowering cognitive modelers to investigate richer, uncharted theoretical spaces.

\section{Methods and Materials}
The studies conducted in the tutorial were not preregistered. All ANNs were implemented in TensorFlow, and the code to reproduce our results is publicly available on GitHub:~\url{https://github.com/ti55987/DL4CM}.  

\textbf{Generative artificial intelligence (Gen AI) disclosure}: Gemini \parencite{google2025gemini} was used solely for language editing to improve the clarity, fluency, and grammar of the manuscript (e.g., using prompts such as "Please revise the following paragraph to be more fluent and clear, without grammatical errors"). Claude \parencite{anthropic2026} was used for refactoring public code and adding comments to functions. All Gen AI output was inspected and validated by the authors. The authors take full responsibility for the text and code.

\section*{Acknowledgments}
This work was funded by the National Institutes of Health (NIH R21MH132974 to AGEC). T.P. is supported by the Berkeley Graduate Fellowship. We are grateful to Xijian Deng, Ijin Yu, and Gaia Molinaro for their validation of the proposed workflow.

\printbibliography

\appendix

\section{Experiment details}
\label{sup:exp_details}
\subsection{Working Memory Model}
Similar to the RL-F model, the working memory (WM) model tracks stimulus-action associations $W(s,a)$ using the standard delta rule \parencite{sutton1998reinforcement}. At trial $t$, after observing the stimulus, action, and reward $(s_t, a_t, r_t)$, the model updates the selected association. All association strengths are initialized to $W_0 = \frac{1}{n_A}$ and updated as follows:
{\small
\[
\begin{aligned}
W_{t+1}(s_t, a_t) = W_t(s_t, a_t) + \alpha_{\text{WM}}(r_t)\left(r_t - W_t(s_t, a_t)\right),  
\text{where } \alpha_{\text{WM}}(r_t) = 
\begin{cases} 
1, & r_t = 1, \\[-6pt]
\text{bias}_{\text{WM}}, &  r_t = 0.
\end{cases}
\end{aligned}
\]
}
where the high learning rate $\alpha_{\text{WM}}$  models one-shot information storage, while using a $\text{bias}_{\text{WM}}$ parameter to capture a potential neglect of negative outcomes ($r_t=0$; \cite{van2012striatum}. It then incorporates memory decay toward the initial value:
\[
\forall (s, a), W_{t+1}(s, a) = W_t(s, a) + \phi_{\text{WM}} \left(W_0 - W_t(s, a)\right)
\]
where $\phi_{\text{WM}}$ is a decay rate parameter. The WM produces the final action using the same softmax policy as the RLF model, incorporating a sticky parameter $\kappa$ to capture choice perseveration. Finally, a random lapse in choices is added to the policy $\pi_{\text{WM}}$ with a noise parameter $\epsilon$:
\[
\pi_{\text{WM}} \leftarrow (1 - \epsilon)\pi_{\text{WM}} + \epsilon \frac{1}{n_A}
\]
To ensure reliable parameter recovery, we fixed the inverse temperature parameter $\beta$ at 25 \parencite{wilson2019ten}. The WM model thus has four free parameters: $\text{bias}_{\text{WM}} \in [0, 1]$, $\phi_{\text{WM}} \in [0, 1]$, $\kappa \in [-1, 1]$, and $\epsilon \in [0, 1]$.

\subsection{Evaluation Metrics}
For all the metrics in this work, results were averaged across agents, where markers and error bars represent the mean and 95\% CI over all test agents.
\label{sup:metrics}
\paragraph{RMSE for Model Parameters} RMSE measures the difference between true ($\theta_c$) and predicted continuous parameters ($\hat{\theta_c}$) from the estimators. RMSE per parameter is defined as:
\begin{equation}
\text{RMSE}\,(\mathbf{\theta_c},\mathbf{\hat{\theta_c}})=  \sqrt{(\theta_i - \hat{\theta}_i)^2}    
\end{equation} 
\paragraph{RMSE for Trial-varying Variables} RMSE measures the difference between true ($Z_c$) and predicted continuous latent variables ($\hat{Z_c}$) from the estimators. RMSE across trials 1 to $T$ is defined as:
\begin{equation}
\text{RMSE}\,(\mathbf{Z_c},\mathbf{\hat{Z_c}})=  \sqrt{\frac{1}{T}\sum_{i=1}^{T}(z_i - \hat{z}_i)^2}    
\end{equation}

\paragraph{Accuracy} We quantified model classification performance using accuracy. For each test agent, the candidate model with the highest predicted probability by the neural estimator was selected as the predicted model identity, and accuracy was calculated as the fraction of instances where the predicted identity matched the ground-truth generating model.

\paragraph{Feature Importance} We quantified feature importance using global permutation tests on the trained neural estimator. First, we computed the baseline loss on the test dataset with an unpermuted feature set. For each feature, we independently shuffled its temporal values within each sample (i.e., agent), then recomputed the loss. Feature importance was defined as the increase in loss relative to baseline, averaged across three independent permutations. Larger increases indicate that the model relies more strongly on the feature for accurate predictions.

\subsection{Details of Artificial Neural Networks}
\paragraph{Architecture} For model parameter estimation and model comparison, we adopted the same network structure as \cite{rmus2024artificial}, which consists of a single Gated Recurrent Unit (GRU) layer followed by three multilayer perceptron (MLP) layers. To estimate model parameters, the final output layer employed a linear activation function and was optimized using a mean squared error (MSE) loss. For model comparison, we used a softmax activation function with a categorical cross-entropy loss.  We applied the rectified linear unit (ReLU) activation function across all MLP layers.

To infer trial-varying latent variables, we adopted LaseNet \parencite{pan2025latent}, which features a bidirectional recurrent neural network (bi-RNN) layer followed by three MLP layers. In this architecture, the number of units in each successive MLP layer is halved. ReLU activation function is applied across all MLP layers, culminating in a linear activation function at the final output layer with an MSE loss to infer continuous $Q$-values in an RLF model.

\paragraph{Neural Networks Training}  The training was run on Nvidia L4 or A10 GPUs, each equipped with 25 GB of memory, and required at most 30 minutes to complete (Appendix Table~\ref{tab:feature_ablation}). Network parameters were randomly initialized and optimized by Adam, with a learning rate of \(3*10^{-4}\). Each network was trained with at most 600 epochs and a batch size of 128. To avoid overfitting, we used 35 or 50 epochs as our early stopping criteria based on validation loss. Hyperparameters were fine-tuned using Bayesian optimization algorithms applied to the validation set (Appendix Table~\ref{table:hyper}; \cite{bergstra2013making}). We allocated 10\% of the training data as the validation set in fine-tuning.

\begin{table}[b!]
\caption{Recommended feature pre-processing}
  \label{app:feature}
\centering
\small
\renewcommand{\arraystretch}{0.5}
\begin{tabular}{lll}
\toprule
\textbf{Feature Type} & \textbf{Example} & \textbf{Recommended Representation} \\
\midrule
Continuous & model parameter $\beta$ & Standardize \\
Large positive continuous & RT = 200--3000 & Standardize; possibly log-transform \\
Binary & reward = 0/1 & 0/1 \\
Nominal category & model identity & One-hot \\
Ordinal & difficulty = low/medium/high & \makecell{Ordinal encoding if ordering is meaningful; \\ otherwise one-hot} \\
\bottomrule
\end{tabular}
\end{table}

\newcommand{\valpm}[2]{\makecell{$#1$ \\[-4pt] \scriptsize{$\pm #2$}}}
\newcommand{\valpmbold}[2]{\makecell{$\bm{#1}$ \\[-4pt] \scriptsize{$\bm{\pm #2}$}}}

\begin{table}[t!]
\centering
\small
\renewcommand{\arraystretch}{0.2}
\caption{\textbf{Feature ablation variants for RLF parameter recovery:} parameter estimation errors and training duration. Values report the mean $\pm$ standard deviation across 3 random seeds. The base feature set comprises actions, rewards, previous actions, delay since last stimulus, \texttt{stim\_iter}, and \texttt{is\_switch}. Bold entries denote the outstanding performance degradation (higher error).}
\label{tab:feature_ablation}
\begin{tabular}{lcccccc}
\toprule
\textbf{Feature Set} & \multicolumn{4}{c}{\textbf{Parameter Error (RMSE)}} & \textbf{Overall} & \textbf{Time} \\
\cmidrule(lr){2-5}
& $\alpha$ & $\beta$ & $\phi$ & $\kappa$ & \textbf{RMSE} & \textbf{(min)} \\
\midrule
Base                            & \valpm{0.122}{0.003}     & \valpm{0.120}{0.003} & \valpm{0.036}{0.002} & \valpm{0.126}{0.002}     & \valpm{0.101}{0.002}     & \valpm{6.8}{0.1} \\
Replace \texttt{is\_switch} with block number     & \valpm{0.117}{0.003}     & \valpm{0.128}{0.001} & \valpm{0.037}{0.001} & \valpm{0.133}{0.009}     & \valpm{0.104}{0.002}     & \valpm{6.7}{0.2} \\
Replace \texttt{is\_switch} with trial number     & \valpm{0.127}{0.004}     & \valpm{0.126}{0.004} & \valpm{0.039}{0.002} & \valpm{0.137}{0.008}     & \valpm{0.107}{0.002}     & \valpm{7.6}{0.7} \\
Remove \texttt{is\_switch}                       & \valpm{0.118}{0.003}     & \valpm{0.123}{0.002} & \valpm{0.037}{0.001} & \valpm{0.127}{0.001}     & \valpm{0.101}{0.001}     & \valpm{6.6}{0.3} \\
Remove \texttt{prev\_action}                     & \valpm{0.131}{0.001}     & \valpm{0.138}{0.001} & \valpm{0.042}{0.001} & \valpmbold{0.178}{0.004} & \valpmbold{0.122}{0.001} & \valpm{6.2}{0.3} \\
Replace \texttt{stim\_iter} with unsorted stimuli & \valpmbold{0.149}{0.004} & \valpm{0.131}{0.002} & \valpm{0.038}{0.001} & \valpm{0.138}{0.004}     & \valpm{0.114}{0.002}     & \valpm{6.4}{0.3} \\
\bottomrule
\end{tabular}
\end{table}

\begin{table}[b!]
  \vspace{-25pt}
  \small
  \renewcommand{\arraystretch}{0.2}    
  \caption{Hyperparameters selection with Bayesian optimization algorithms}
  \label{table:hyper}
  \begin{tabular}{lll}
    \toprule
    \cmidrule(r){1-2}
    Hyperparameter    & Sweep range      \\
    \midrule
    \# of units in RNN layer  &  $96, 128, 512$ \\
    dropout rate in RNN layer &   $\mathcal{U}(0.015, 0.25)$ \\
    dropout rate in 1st MLP layer  & $\mathcal{U}(0.01, 0.2)$ &  \\
    dropout rate in 2nd MLP layer & $\mathcal{U}(0.01, 0.2)$  \\
    learning rate &  $\text{Log-}\mathcal{U}(0.003, 0.03)$ \\
    batch size &  $128, 256, 512$ \\
    \bottomrule   
  \end{tabular}
\end{table}

\subsection{Details of Likelihood-dependent Methods}
For model parameter estimation, we inferred the best-fitting model parameters $\theta$ given observable data \textbf{Y} by either maximizing a likelihood function $P(\mathbf{Y}\mid\theta)$ or the posterior probability $P(\mathbf{Y}\mid\theta)\,\,P(\theta)$.  To infer trial-varying latent variables, we employed a two-step fitting procedure. First, we used MLE or MAP to find the best-fitting model parameters $\tilde{\theta}$. Second, we derived the latent variables \textbf{Z}, we approximated them from the targeted cognitive models \textbf{f} with the obtained best-fitting parameters $\tilde{\theta}$ and \textbf{Y}:
\begin{math}
\mathbf{\tilde{Z}} \approx  \mathbf{f}(\mathbf{Y};\tilde{\theta})
\end{math}.
Model identification was based on AIC metrics \parencite{akaike1974new}. All model fitting was implemented using Python’s \href{https://scipy.org/}{scipy} library. The L-BFGS-B algorithm was applied for bound-constrained minimization using eight random starting points and a maximum of 200 iterations per agent \parencite{byrd1995limited}. 

\section{Parameter Recovery for Models with Intractable Likelihood}
\label{sup:intractable_example}
\paragraph{Cognitive model and task}
To demonstrate the use case of neural SBI on models with intractable likelihood, we adopted the noisy reinforcement learning (Noisy RL) model from \cite{findling2019computational}. Noisy RL is a simple but intractable model performing a two-armed bandit task with complete feedback (Experiment 3 in \cite{findling2019computational}). In this task, agents choose between two options and observe the continuously drifting payoffs of both the obtained $r_t(a)$ and foregone rewards. The model captures behavior using a Rescorla-Wagner update rule with separate learning rates for chosen ($\alpha_c$) and unchosen ($\alpha_u$) actions: 
\noindent{}
\begin{equation}
\begin{gathered}
\delta_t(a) = r_t(a) - Q_t(a) \\
Q_{t+1}(a) = Q_t(a) + \alpha_a * \delta_t(a) + \eta_t(a), \qquad \eta_t(a) \sim \mathcal{N}\left(0, (\epsilon\,|\delta_t(a)|)^2\right)
\end{gathered}
\end{equation}
\noindent{}
where $\alpha_a$ is $\alpha_c$ if $a$ was chosen on trial $t$ and $\alpha_u$ otherwise. Crucially, each Q-value $Q_t(a)$ update is corrupted by Gaussian noise $\eta_t(a)$ proportional to the prediction error $\delta_t(a)$ (following Weber's law), governed by a learning noise $\epsilon$. Because this stochasticity accumulates across trials within the unobserved latent action values $Q(a)$, computing the exact likelihood of a choice sequence requires integrating over all possible latent value trajectories, rendering the model likelihood intractable (Fig~\ref{fig:sim_intractable_exp}D).

Like the RLF Model, the noisy RL uses a softmax choice policy to select actions, controlled by an inverse temperature $\beta$ and a repetition bias $\kappa$. In our experiment, we chose a model with fixed $\beta=35$ because it captures human choices the best under complete feedback \parencite{findling2019computational}. The noisy RL thus includes four free parameters: $\alpha_c$, $\alpha_u$, $\epsilon$, and $\kappa$.

\paragraph{Neural SBI workflow} We simulated two training sets of 3,000 and 9,000 agents (with a 10\% validation split) and a separate test set of 500 agents. Each agent completed the 896-trial task (16 blocks of 56 trials). For each agent, we sampled the four target unobservable parameters from uniform priors: $\alpha_c \in [0.05, 0.8]$, $\alpha_u \in [0.05, 0.8]$, $\epsilon \in [0.01, 1]$, and $\kappa \in [-0.25, 0.5]$. These model parameters were the targets of inference for real participant data.

Before training, we extracted nine observable features (\texttt{actions}, \texttt{prev\_action}, \texttt{reward1\_bin}, \texttt{reward0\_bin}, \texttt{chosen\_reward\_bin}, \texttt{unchosen\_reward\_bin}, \texttt{chosen\_vs\_unchosen\_reward\_bin}, \texttt{is\_switch}, and \texttt{optimal\_choice}; Fig.~\ref{fig:features_process}). Notably, because task rewards are continuous on $[0, 1]$, we discretized them into five equal-width bins over $[0,1]$ (i.e., \texttt{reward1\_bin}, \texttt{reward0\_bin}, \texttt{chosen\_reward\_bin}, \texttt{unchosen\_reward\_bin}), and the difference between chosen and unchosen rewards into five equal-width bins over $[-1, 1]$ (\texttt{chosen\_vs\_unchosen\_reward\_bin}). This discretization significantly reduces the number of possible values of the continuous feature, avoiding the slow and inefficient training often caused by high-cardinality continuous inputs \parencite{dougherty1995supervised, liu2002discretization}. 

\begin{figure}[t!]
    \centering
    \includegraphics[width=1\linewidth]{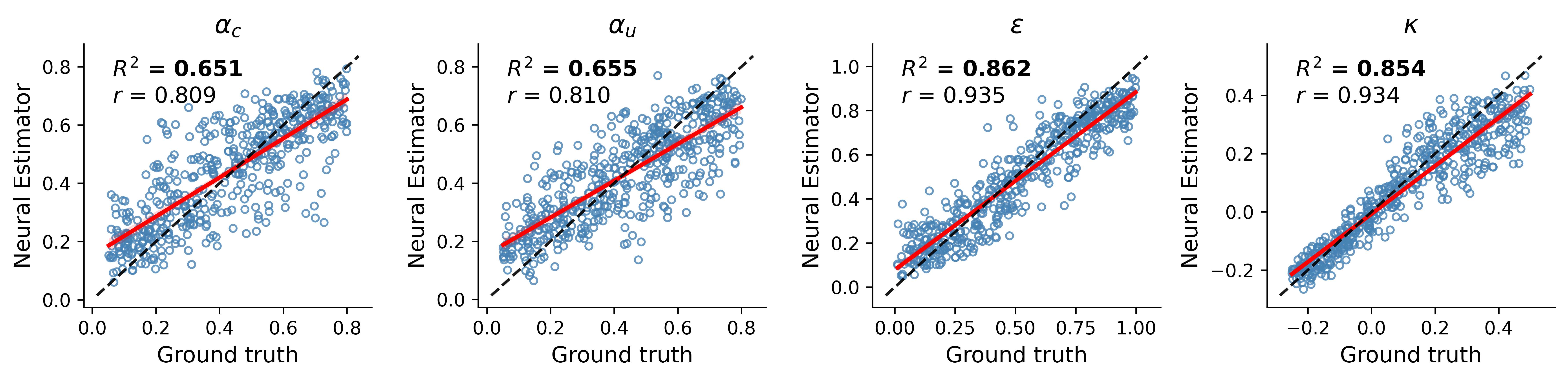}  
    \caption{Noisy RL parameter recovery: recovery results from 500 test agents. The model was trained on a separate dataset comprising 9000 agents.}    
    \label{sub_fig:noisy_rl_pe}
    \vspace{-25pt}    
\end{figure}

We first trained a neural estimator on all nine features using the 3,000-agent dataset. Evaluating feature importance via permutation tests identified the four most informative features: \texttt{actions}, \texttt{prev\_action}, \texttt{reward1\_bin}, and \texttt{chosen\_vs\_unchosen\_reward\_bin}. To boost inference performance, we then trained our final neural estimator using only this four-feature set on the larger 9,000-agent dataset (Fig.~\ref{sub_fig:noisy_rl_pe}).

\paragraph{Validation of feature importance} Table~\ref{tab:noise_rl_addition} summarizes parameter recovery on the 500 test agents for estimators trained on 3,000 agents. The four selected features achieved an overall RMSE of 0.130, and adding any of the other five features did not improve recovery. This suggests that the four features capture sufficient information about the model dynamics to infer the target parameters.

The limited contribution of the remaining features stems from two factors: informational redundancy and constrained task structure. First, \texttt{reward1\_bin} and \texttt{chosen\_vs\_unchosen\_reward\_bin} jointly encode the relevant reward information, making \texttt{reward0\_bin}, \texttt{chosen\_reward\_bin}, and \texttt{unchosen\_reward\_bin} redundant. Second, \texttt{is\_switch} and \texttt{optimal\_choice} are determined by the task structure. Because our training dataset was simulated the same block structure, the neural estimator can implicitly learn these structural regularities from the thousands of behavioral trajectories.   Consistent with the RLF example in the main text, pruning redundant or deterministic task features reduces input dimensionality and modestly improves parameter recovery at a fixed simulation budget.

\begin{table}[htbp]
\centering
\small
\renewcommand{\arraystretch}{0.2}
\caption{\textbf{Noisy RL parameter recovery:} Parameter estimation errors, overall recovery performance, and training duration across incremental feature additions. The base feature set comprises actions, previous actions, \texttt{reward1\_bin}, and \texttt{chosen\_vs\_unchosen\_reward\_bin}. Values report the mean $\pm$ standard deviation across 3 random seeds. Bold entries denote the lowest estimation error for each parameter and overall RMSE.}
\label{tab:noise_rl_addition}
\begin{tabular}{lcccccc}
\toprule
\textbf{Feature Set} & \multicolumn{4}{c}{\textbf{Parameter Error (RMSE)}} & \textbf{Overall} & \textbf{Time} \\
\cmidrule(lr){2-5}
& $\alpha_c$ & $\alpha_u$ & $\epsilon$ & $\kappa$ & \textbf{RMSE} & \textbf{(min)} \\
\midrule
Base (4)                             & \valpmbold{0.168}{0.018} & \valpmbold{0.145}{0.008} & \valpm{0.128}{0.002}     & \valpm{0.079}{0.002}     & \valpmbold{0.130}{0.006} & \valpm{1.9}{0.0} \\
+ \texttt{reward0\_bin} (5)          & \valpm{0.178}{0.014}     & \valpm{0.160}{0.021}     & \valpm{0.119}{0.002}     & \valpm{0.077}{0.001}     & \valpm{0.133}{0.008}     & \valpm{2.2}{0.7} \\
+ \texttt{chosen\_reward\_bin} (6)   & \valpm{0.169}{0.020}     & \valpm{0.159}{0.027}     & \valpm{0.120}{0.007}     & \valpmbold{0.076}{0.003} & \valpm{0.131}{0.012}     & \valpm{1.9}{0.6} \\
+ \texttt{unchosen\_reward\_bin} (7) & \valpm{0.190}{0.001}     & \valpm{0.191}{0.001}     & \valpmbold{0.117}{0.003} & \valpm{0.077}{0.001}     & \valpm{0.144}{0.001}     & \valpmbold{1.3}{0.1} \\
+ \texttt{is\_switch} (8)          & \valpm{0.171}{0.017}     & \valpm{0.183}{0.002}     & \valpm{0.120}{0.002}     & \valpm{0.078}{0.001}     & \valpm{0.138}{0.005}     & \valpm{2.2}{0.2} \\
+ \texttt{optimal\_choice} (9)       & \valpm{0.171}{0.009}     & \valpm{0.167}{0.008}     & \valpm{0.120}{0.008}     & \valpm{0.081}{0.001}     & \valpm{0.135}{0.006}     & \valpm{2.5}{0.9} \\
\bottomrule
\end{tabular}
\end{table}

\end{document}